\documentclass[letterpaper, 10 pt, conference]{ieeeconf}  % Comment this line out if you need a4paper

\IEEEoverridecommandlockouts                              % This command is only needed if 
\usepackage{graphics} % for pdf, bitmapped graphics files
\usepackage{epsfig} % for postscript graphics files
\usepackage{amsmath} % assumes amsmath package installed
\usepackage{amssymb}  % assumes amsmath package installed
\usepackage{xcolor}
\usepackage{multirow}
\usepackage{booktabs}
\usepackage{hyperref}
\usepackage{cleveref}
\usepackage{pifont}
\usepackage{utfsym}
\usepackage{setspace} 
\usepackage{algorithm}
\usepackage{makecell}
\usepackage[table]{xcolor}
\usepackage{cite}

\title{\LARGE \bf
UniUncer: Unified Dynamic–Static Uncertainty for End-to-End Driving
}

\author{Yu Gao$^{1\dagger*}$, Jijun Wang$^{1,2*\ddagger}$, Zongzheng Zhang$^{2}$, Anqing Jiang$^{1}$, Yiru Wang$^{1}$, \\
Yuwen Heng$^{1}$, Shuo Wang$^{1}$, HAO SUN$^{1}$, Zhangfeng Hu$^{3}$, Hao Zhao$^{2}$% <-this % stops a space
\thanks{$^{1}$Bosch Corporate Research, China.}%
\thanks{$^{2}$Institute for AI Industry Research (AIR), Tsinghua University, China.}%
\thanks{$^{3}$Rensselaer Polytechnic Institute, USA.}%
\thanks{$^*$Equal contribution.}
\thanks{$^\dagger$Corresponding author.}
\thanks{$^\ddagger$Work done during the internship at Bosch Corporate Research, China.}
}

\begin{document}

\maketitle
\thispagestyle{empty}
\pagestyle{empty}

%%%%%%%%%%%%%%%%%%%%%%%%%%%%%%%%%%%%%%%%%%%%%%%%%%%%%%%%%%%%%%%%%%%%%%%%%%%%%%%%

\begin{abstract}
End-to-end (E2E) driving has become a cornerstone of both industry deployment and academic research, offering a single learnable pipeline that maps multi-sensor inputs to actions while avoiding hand-engineered modules. However, the reliability of such pipelines strongly depends on how well they handle uncertainty: sensors are noisy, semantics can be ambiguous, and interaction with other road users is inherently stochastic. Uncertainty also appears in multiple forms: classification vs. localization, and, crucially, in both static map elements and dynamic agents. Existing E2E approaches model only static-map uncertainty, leaving planning vulnerable to overconfident and unreliable inputs. We present UniUncer, the first lightweight, unified uncertainty framework that jointly estimates and uses uncertainty for both static and dynamic scene elements inside an E2E planner. Concretely: (1) we convert deterministic heads to probabilistic Laplace regressors that output per-vertex location and scale for vectorized static and dynamic entities; (2) we introduce an uncertainty-fusion module that encodes these parameters and injects them into object/map queries to form uncertainty-aware queries; and (3) we design an uncertainty-aware gate that adaptively modulates reliance on historical inputs (ego status or temporal perception queries) based on current uncertainty levels. The design adds minimal overhead and drops throughput by only $\sim$0.5 FPS while remaining plug-and-play for common E2E backbones. On nuScenes (open-loop), UniUncer reduces average L2 trajectory error by 7\%. On NavsimV2 (pseudo closed-loop), it improves overall EPDMS by 10.8\% and notable stage two gains in challenging, interaction-heavy scenes. Ablations confirm that dynamic-agent uncertainty and the uncertainty-aware gate are both necessary. Qualitatively, UniUncer produces well-calibrated uncertainties that encourage human-like decisions when evidence is unreliable, improving robustness without sacrificing efficiency. Our project page is: https://bradgers.github.io/Uniuncer.

\end{abstract}

\section{Introduction}

End-to-end (E2E) autonomous driving models~\cite{hu2022st, wu2022trajectory, tong2023scene, jia2023think ,hu2023planning,jiang2023vad,yan2023int2,tian2023unsupervised,jia2023driveadapter,jin2024tod3cap,li2024ego,chen2024ppad,zheng2024genad,weng2024drive,sun2025sparsedrive,liao2025diffusiondrive,yang2025uncad,xing2025goalflow,zhang2025bridging, song2025don,li2024enhancing, li2024navigation, jia2025drivetransformer, li2025avd2,zhang2025chameleon,ding2025hint, li2025reusing,chi2025impromptu, zheng2025world4drive, lu2025real, fu2025orion, tang2025hip, guo2025dist, li2025hydra, jiang2025diffvla, zhang2025delving} have emerged as a promising paradigm that directly maps sensor inputs to driving decisions. Compared to modular pipelines that separately process perception, prediction and planning, E2E models jointly optimize all stages within a single framework, resulting in simpler architectures and reduced information loss across modules. This holistic approach has made E2E models increasingly attractive for both academic research and large-scale industrial deployment.

\begin{figure}[t]
\centering 
    \includegraphics[width=0.7\linewidth]{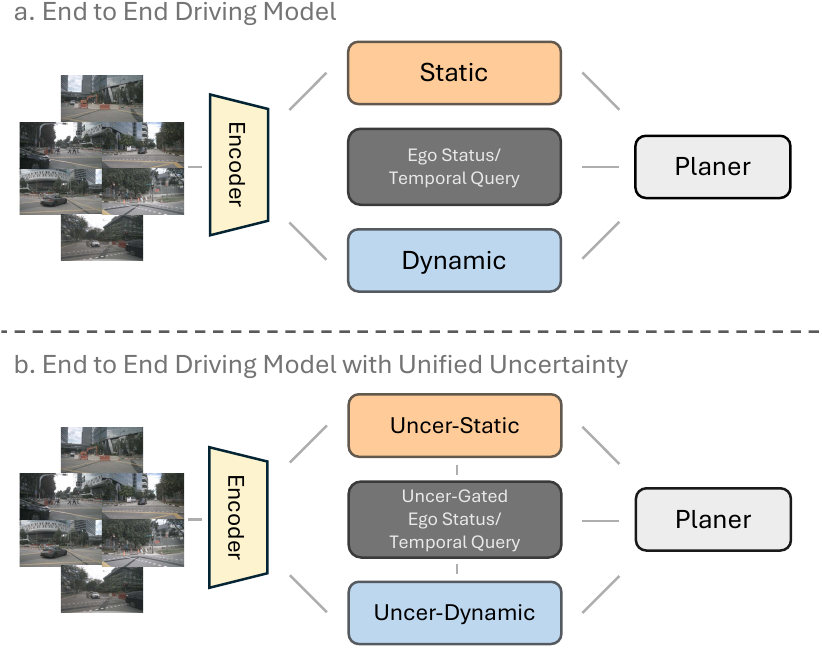}
    \caption{Comparison between (a) a standard E2E autonomous driving model and (b) our uncertainty-aware E2E model. Our model introduces uncertainty estimation in both static and dynamic branches, and employs an uncertainty-aware gating mechanism to modulate historical information based on current uncertainty levels.}
\label{fig:teaser}
\vspace{-15pt}
\end{figure}

Early works primarily investigated the essential components of an E2E system. UniAD~\cite{hu2023planning} integrated perception, prediction, and planning with explicit tracking and mapping modules, while VAD~\cite{jiang2023vad} distilled the design into three core modules: dynamic agents, static map, and planning. Building on these foundations, recent efforts have further expanded the paradigm by incorporating generative modeling~\cite{zheng2024genad,liao2025diffusiondrive,xing2025goalflow}, world models~\cite{li2025end, zheng2025world4drive}, and large language models (LLMs)~\cite{chi2025impromptu,fu2025orion, lu2025real}, thereby enriching scene reasoning, enhancing generalization, and enabling longer-horizon prediction.

Despite these advances, the foundation of E2E systems still lies in the reliable perception of two fundamental elements: static map structures and dynamic agents. However, both elements are inherently uncertain due to noisy sensors, occlusions, ambiguous semantics, and the unpredictable behaviors of surrounding vehicles. While architectural innovations have significantly improved model expressiveness and scalability, explicit modeling of uncertainty, particularly a unified treatment of static and dynamic uncertainties, remains relatively underexplored. This gap is critical, as uncertainty estimation directly influences planning robustness and safety of autonomous systems in complex real-world environments.

\begin{figure*}[t]
\centering 
    \includegraphics[width=0.9\textwidth]{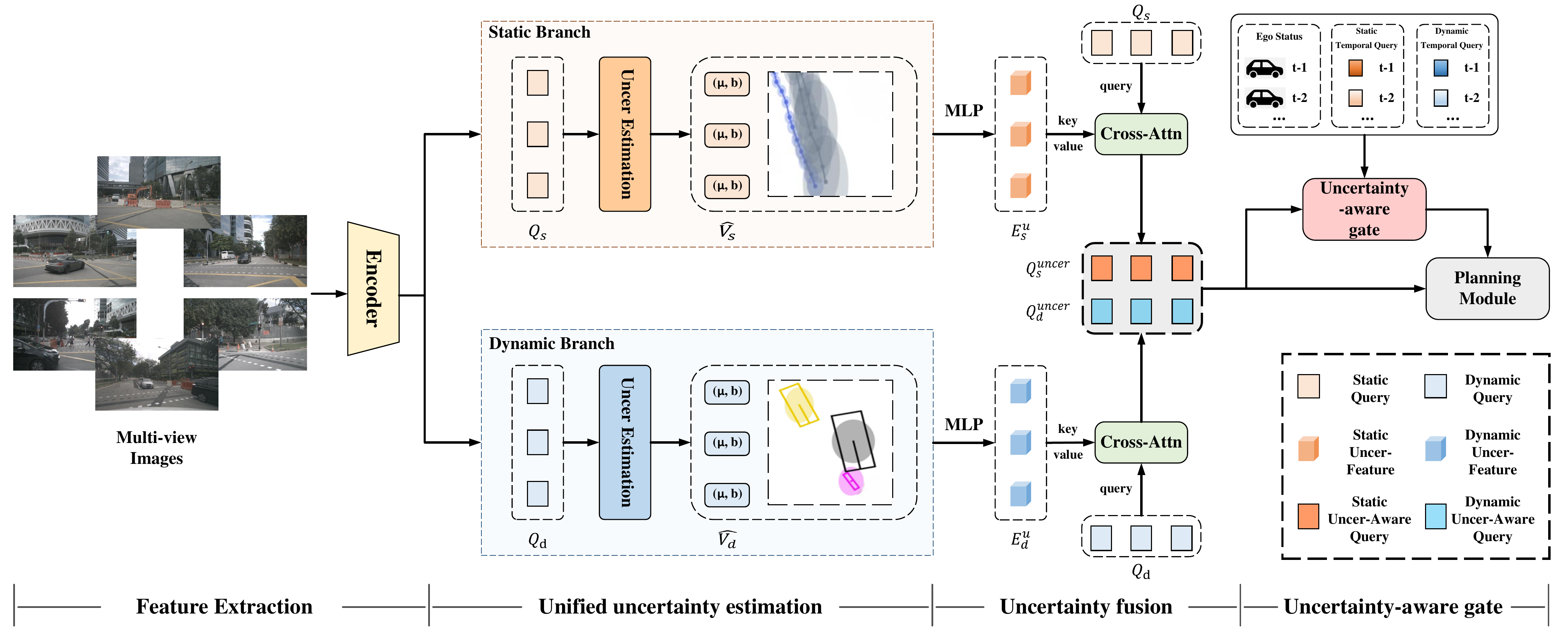} 
    \caption{Overview of the UniUncer framework. UniUncer extends the static and dynamic branches with a unified uncertainty estimation head to predict Laplace parameters. These parameters are encoded and fused with the original queries to form uncertainty-aware queries, which serve as input to the planner and gate historical information (e.g., ego state and temporal queries) via an uncertainty-aware gate.} 
\label{fig:overview} 
\vspace{-15pt}
\end{figure*}

Uncertainty has been studied extensively in modular pipelines, particularly in online mapping~\cite{gu2024producing,zhang2025delving}, and UncAD~\cite{yang2025uncad} recently extended map-related uncertainty into E2E systems. However, improvements achieved by modeling map uncertainty alone in E2E settings are limited. To fully exploit uncertainty in E2E systems requires, on the one hand, a unified framework that jointly models both dynamic and static uncertainties, and on the other hand, an effective mechanism to incorporate these uncertainties into planning. Nevertheless, uncertainty as a core component remains largely overlooked, and the potential benefits of unified uncertainty modeling for improving E2E performance are still not well understood. Without explicitly accounting for uncertainties, planning modules may over-rely on unreliable inputs, resulting in degraded robustness and compromised safety in safety-critical driving scenarios.

As illustrated in Fig.~\ref{fig:teaser}, a standard E2E model processes all perception features without considering their uncertainty, and integrates all historical information (e.g., past ego status or temporal perception queries) into the planner indiscriminately. In contrast, our uncertainty-aware E2E model in Fig~\ref{fig:teaser}, b explicitly estimates both static and dynamic uncertainties, and uses this information to adaptively gate historical inputs.

In this work, we revisit the role of uncertainty in E2E autonomous driving and propose lightweight modules that explicitly estimate and exploit static and dynamic uncertainties. As shown in Fig.~\ref{fig:overview}, our framework, named UniUncer, enhances planning robustness with minimal architectural modifications and negligible computational overhead. Our main contributions are as follows:
\begin{itemize}
    \item \textbf{Unified uncertainty estimation:} We extend deterministic regression heads in both static and dynamic branches to probabilistic ones, enabling uncertainty-aware feature representations with only minor changes to the existing design.
    \item \textbf{Uncertainty fusion:} We introduce a module that systematically integrates static and dynamic uncertainties into object features, improving downstream planning decisions in challenging scenarios.
    \item \textbf{Uncertainty-aware gate:} We design an adaptive gating mechanism that modulates reliance on historical ego status and temporal perception queries depending on current uncertainty levels, thereby improving the use of historical information for planning decisions.
\end{itemize}

Extensive experiments on both open-loop (nuScenes) and pseudo closed-loop (NavsimV2) benchmarks demonstrate the effectiveness of our approach. Our method achieves more than $7\%$ reduction in L2 error on nuScenes and a $10\%$ improvement in EPDMS on NavsimV2, highlighting the benefits of uncertainty-aware modeling for improving the reliability of E2E autonomous driving.

\section{Related Works}

\subsection{End-to-End Driving Models}
End-to-end (E2E) autonomous driving aims to directly map raw sensor inputs to ego vehicle planning decisions, bypassing the explicit decomposition into modular perception–prediction–planning pipelines. Early approaches such as Transfuser~\cite{chitta2022transfuser} introduced attention-based fusion for multi-modal inputs, while UniAD~\cite{hu2023planning} pioneered a differentiable unified framework that integrates perception, prediction, and planning in a joint optimization process. Subsequent designs have focused on improving efficiency and scene representation: VAD~\cite{jiang2023vad} adopted vectorized scene modeling, Para-Drive~\cite{weng2024drive} proposed a parallelized architecture, and SparseDrive~\cite{sun2025sparsedrive} introduced a fully sparse design that scales favorably with scene complexity. More recently, generative paradigms~\cite{zheng2024genad,liao2025diffusiondrive,xing2025goalflow}, world-model-based approaches~\cite{li2025end, zheng2025world4drive}, and large language foundation models~\cite{chi2025impromptu,fu2025orion, lu2025real} have been explored to enhance reasoning, generalization, and long-horizon planning.

Despite these architectural and representational advances, most E2E models treat perception outputs as deterministic and fully reliable, overlooking the inherent uncertainty in both static and dynamic scene elements. While some efforts have begun to incorporate uncertainty, particularly in static map modeling~\cite{yang2025uncad}, the unified treatment of dynamic and static uncertainties, and their explicit integration into planning, remains largely unaddressed. In contrast, UniUncer introduces a lightweight yet effective framework that jointly estimates and leverages both types of uncertainty to improve planning robustness and adaptability.

\begin{table*}[htbp]
% \vspace{+5pt}
\caption{Open-loop E2E performance benchmark on nuScenes.}
\centering
\begin{tabular}{l | c | c | c | c c c c | c c c c}
\toprule
\multirow{2}{*}{Method} & \multirow{2}{*}{Reference} & \multirow{2}{*}{Input} & \multirow{2}{*}{Img. Backbone} & \multicolumn{4}{c|}{L2 (m) $\downarrow$} & \multicolumn{4}{c}{Col. Rate (\%) $\downarrow$} \\
\cline{5-12}
 &  &  & & 1s & 2s & 3s & Avg. & 1s & 2s & 3s & Avg. \\
\midrule
ST-P3~\cite{hu2022st} & ECCV 2022 & Camera & EffNet-b4 & 1.33 & 2.11 & 2.90 & 2.11 & 0.23 & 0.62 & 1.27 & 0.71 \\
OccNet~\cite{tong2023scene} &  ICCV 2023 & Camera & ResNet-50 & 0.45 & 0.70 & 1.04 & 0.73 & 0.62 & 0.58 & 0.63 & 0.61 \\
UniAD~\cite{hu2023planning} & CVPR 2023 & Camera & ResNet-101 & 0.45 & 0.70 & 1.04 & 0.73 & 0.62 & 0.58 & 0.63 & 0.61 \\
VAD~\cite{jiang2023vad} & ICCV 2023 & Camera & ResNet-50 & 0.41 & 0.70 & 1.05 & 0.72 & 0.03 & 0.19 & 0.43 & 0.21 \\
GenAD~\cite{zheng2024genad} &  ECCV 2024 & Camera & ResNet-50 & 0.36 & 0.83 & 1.55 & 0.91 & 0.06 & 0.23 & 1.00 & 0.43 \\
DiffusionDrive~\cite{liao2025diffusiondrive} & CVPR 2025 & Camera & ResNet-50 & 0.27 & 0.54 & 0.90 & 0.57 & 0.03 & 0.05 & \textbf{0.16} & 0.08 \\
BridgeAD~\cite{zhang2025bridging} & CVPR 2025 & Camera & ResNet-50 & 0.29 & 0.57 & 0.92  & 0.59 & 0.01 & 0.05 & 0.22 & 0.09 \\
\midrule[1.2pt]
SparseDrive~\cite{sun2025sparsedrive} & ICRA 2025 & Camera & ResNet-50 & 0.29 & 0.58 & 0.96 & 0.61 & 0.01 & 0.05 & 0.18 & 0.08 \\
UncAD~\cite{yang2025uncad} & ICRA 2025 & Camera & ResNet-50 & 0.29 & 0.57 & 0.95 & 0.60 & \textbf{0.00} &\textbf{0.04} & 0.18 & \textbf{0.07} \\
% SparseDrive + Ours & - & Camera & ResNet-50 & \textbf{0.26} & \textbf{0.50} & \textbf{0.85} & \textbf{0.54} & \textbf{0.00} & 0.05 & 0.18 & 0.08 \\
SparseDrive + Ours & - & Camera & ResNet-50 & \textbf{0.27} & \textbf{0.53} & \textbf{0.90} & \textbf{0.57} & \textbf{0.00} & \textbf{0.04} & 0.17 & \textbf{0.07} \\
\bottomrule
\end{tabular}
\label{tab:nuscenes_results}
\vspace{-10pt}
\end{table*}

% UncerMask achieves superior performance in terms of L2 error compared to both prior uncertainty-aware and non-uncertainty E2E models. In terms of collision rate, our method maintains comparable performance to previous uncertainty-aware approaches, while consistently outperforming non-uncertainty E2E baselines.

\subsection{Evaluation Protocols}
Evaluating E2E driving models typically involves open-loop, closed-loop, or hybrid settings. Open-loop evaluation, widely used on benchmarks such as nuScenes~\cite{caesar2020nuscenes} and Navsim~\cite{dauner2024navsim}, assesses trajectory accuracy and rule-based safety metrics. However, it fails to capture long-term feedback loops and multi-agent interactions, limiting its realism. Closed-loop evaluation in simulators like CARLA~\cite{dosovitskiy2017carla} and Bench2Drive~\cite{jia2024bench2drive} provides more realistic assessment of driving policies but suffers from computational inefficiency and sim-to-real gaps.

To bridge this divide, NavsimV2~\cite{cao2025pseudo} introduced a pseudo-simulation framework based on 3D Gaussian Splatting (3DGS)~\cite{kerbl20233d}, offering a compelling middle ground by combining high-fidelity scene rendering with efficient execution. This enables realistic evaluation without the overhead of full simulation. In our experiments, we adopt both the nuScenes (open-loop) and NavsimV2 (pseudo closed-loop) benchmarks to comprehensively validate the effectiveness of uncertainty-aware modeling under diverse evaluation paradigms.

\subsection{Uncertainty Estimation}
Uncertainty estimation plays a critical role in safe and reliable decision-making, particularly in ambiguous or safety-critical environments. In perception tasks, including object detection, segmentation, and depth completion, uncertainty modeling has been shown to improve reliability and interpretability~\cite{kendall2017uncertainties,gast2018lightweight,poggi2020uncertainty,meyer2019lasernet}. For trajectory prediction, incorporating map-related uncertainty has been demonstrated to enhance both accuracy and robustness~\cite{gu2024producing,zhang2025delving}. More recently, UncAD~\cite{yang2025uncad} extended uncertainty modeling to E2E driving by integrating online mapping uncertainty into the planning process.

However, these approaches focus primarily on static map uncertainty and neglect the uncertainty associated with dynamic agents, an equally critical source of risk in interactive driving scenarios. Moreover, they lack a unified mechanism to jointly model and exploit both static and dynamic uncertainties during planning.

\section{Method}
\label{sec:Method} 

This section begins by introducing UniUncer, a unified uncertainty-aware E2E framework that models both static and dynamic elements as vectorized Bird's Eye View (BEV) representations with Laplace-distributed vertices to capture spatial uncertainty. We then describe how a unified uncertainty head estimates these uncertainties (Sec.~\ref{sec:uni_uncer_estimation}), how uncertainty-aware queries are generated via uncertainty fusion (Sec.~\ref{sec:uncer_fusion}), and how the enhanced queries dynamically gate historical information through an uncertainty-aware gate (Sec.~\ref{sec:uncer_gate}). As shown in Fig.~\ref{fig:overview}, the E2E system captures images from surrounding-view cameras and uses an encoder to produce perception-related queries $Q$, including static and dynamic queries $Q_s$ and $Q_d$.

\subsection{Unified Uncertainty Estimation}
\label{sec:uni_uncer_estimation}
We model the aleatoric uncertainty of vertex locations with a Laplace distribution. This choice is motivated by its natural connection to the robust $\ell_1$ loss commonly used in object detection and vectorized mapping. Formally, minimizing the $\ell_1$ loss is equivalent to performing maximum likelihood estimation under a Laplace noise assumption with a fixed scale. By generalizing this to predict both the location $\mu$ and the scale $b$ for each vertex, we obtain a principled and learnable measure of spatial uncertainty while maintaining compatibility with existing regression frameworks.

Uncertainty estimation for static objects has proven to be effective in both modular~\cite{gu2024producing,zhang2025delving} and E2E~\cite{yang2025uncad} autonomous driving systems. These approaches explicitly model aleatoric uncertainty, which characterizes the stochastic noise inherent to the measurement process~\cite{kendall2017uncertainties}. They extend vectorized map representations by modeling each element as a set of $N_s$ vertices in the BEV space, with each vertex parameterized by a Laplace distribution in the $x$ and $y$ directions:
\begin{equation*}
\mathbf{v}_i^s = (\mu_x^{(i)}, b_x^{(i)}, \mu_y^{(i)}, b_y^{(i)}) \in \mathbb{R}^4, \quad i = 1, \dots, N_s,
\end{equation*}
where $\mu$ and $b$ denote the location and scale parameters, respectively. 
The Laplace distribution is a natural choice, as most detection models are trained with regression loss $\ell_1$.

To enable a unified representation of static and dynamic objects, we convert the bounding box labels online during training. The conventional box parameters~$(x,y,z,\text{width},\text{length},\text{height},\text{heading})$ are converted into a vectorized BEV representation with $N_d$ vertices, each modeled as:
\begin{equation*}
\mathbf{v}_j^d = (\mu_x^{(j)}, b_x^{(j)}, \mu_y^{(j)}, b_y^{(j)}) \in \mathbb{R}^4, \quad j=1,\dots,N_d.
\end{equation*}

Static and dynamic queries $Q_s$ and $Q_d$ are processed by probabilistic heads (MLPs) to generate vertex predictions:
\begin{equation}
\hat{V}_s = f_p(Q_s), \quad \hat{V}_d = f_p(Q_d),
\end{equation}
where $f_p$ predicts the Laplace distribution parameters for each vertex. The resulting vertex sets are given by
\begin{equation}
\hat{V}_s = \{\mathbf{v}_i^s\}_{i=1}^{N_s}, \quad \hat{V}_d = \{\mathbf{v}_j^d\}_{j=1}^{N_d},
\end{equation}
with each $\mathbf{v}_i^s$ and $\mathbf{v}_j^d$ parameterizing the location and scale in the BEV plane. To ensure strictly positive scale parameters and numerical stability, we apply a Softplus activation with a small offset (\( \epsilon = 10^{-6} \)) to the raw network outputs of scale parameter.

\begin{table*}[htbp]
% \vspace{+5pt}
\scriptsize
\caption{Pseudo Close Loop E2E Performance Benchmark on the navhard two-stage test from NavsimV2.}
\centering
% \rowcolors{2}{gray!30}{gray!15}
\resizebox{\textwidth}{!} {
\begin{tabular}{l| c|c |c c c c c c c c c c | c}
    \toprule
    Method 
    & Backbone
    & Stage
    & $\text{NC $\uparrow$}$
    & $\text{DAC $\uparrow$}$
    & $\text{DDC $\uparrow$}$
    & $\text{TLC $\uparrow$}$
    & $\text{EP $\uparrow$}$
    & $\text{TTC $\uparrow$}$
    & $\text{LK $\uparrow$}$ 
    & $\text{HC $\uparrow$}$ 
    & $\text{EC $\uparrow$}$
    & $\text{EPDMS Stage $\uparrow$}$
    & $\text{EPDMS $\uparrow$}$ \\
    \midrule
    \multirow{2}{*}{LTF~\cite{chitta2022transfuser}} & ResNet-34 & Stage 1 &
    96.2 & 79.6 & 99.1 & 99.6 & 84.1 & 95.1 & 94.2 & 97.6 & 79.1 & 62.3 & \multirow{2}{*}{23.1}  \\
    & ResNet-34 & Stage 2 &
    77.8 & 70.4 & 84.3 & 98.1 & 85.1 & 75.7 & 45.4 & 95.7 & 76.0 & 37.5 &  \\
    \midrule
    
    \multirow{2}{*}{DiffusionDrive~\cite{liao2025diffusiondrive}} & ResNet-34 & Stage 1 &
    96.3 & 84.7 & 98.7 & 99.3 & 84.3 & 95.3 & 94.4 & 97.8 & 76.4 & \textbf{65.5} & \multirow{2}{*}{25.9} \\
    & ResNet-34 & Stage 2 &
    79.5 & 71.2 & 84.9 & 98.3 & 86.5 & 75.6 & 46.9 & 96.2 & 70.9 & 39.6 & \\
    \midrule
    
    \multirow{2}{*}{DiffusionDrive + Ours} & ResNet-34 & Stage 1 &
    96.8 & 82.9 & 98.8 & 99.3 & 83.6 & 95.8 & 96.0 & 97.6 & 77.3 & 64.1 &  \multirow{2}{*}{\textbf{28.7}}\\
    & ResNet-34 & Stage 2 &
    81.5 & 74.4 & 86.9 & 98.1 & 85.2 & 78.6 & 49.2 & 95.7 & 74.2 & \textbf{43.4} & \\
    
\bottomrule
\end{tabular}}
\label{tab:navhard_results}
\vspace{-10pt}
\end{table*}

\subsection{Uncertainty Fusion Module}
\label{sec:uncer_fusion}

The uncertainty fusion module encodes the predicted vertices into uncertainty features $E^u_s$ and $E^u_d$ via an MLP encoder $\text{Uncer}_{\text{enc}}$, given by 
\begin{equation}
\begin{split}
E^u_s &= \text{Uncer}_{\text{enc}}(\hat{V}_s) \in \mathbb{R}^{N_s \times d_h}, \\
E^u_d &= \text{Uncer}_{\text{enc}}(\hat{V}_d) \in \mathbb{R}^{N_d \times d_h},
\end{split}
\end{equation}
where $d_h$ is the hidden dimension, consistent with the query embedding dimension.  

Given queries $Q_s \in \mathbb{R}^{N_s \times d_h}$ and $Q_d \in \mathbb{R}^{N_d \times d_h}$, we refine them by attending to the uncertainty features via multi-head cross-attention:
\begin{equation}
\begin{split}
Q_s^{\text{uncer}} &= \text{Attention}_s(Q_s, E^u_s, E^u_s), \\
Q_d^{\text{uncer}} &= \text{Attention}_d(Q_d, E^u_d, E^u_d),
\end{split}
\end{equation}
producing uncertainty-aware queries $Q_s^{\text{uncer}}$ and $Q_d^{\text{uncer}}$ that replace $Q_s$ and $Q_d$ in downstream planning.

\subsection{Uncertainty-aware Gate}
\label{sec:uncer_gate}

The uncertainty-aware gate modulates historical information based on current scene uncertainty. Depending on the different AD approaches, it generates gating signals at different granularities: for high-dimensional temporal queries, it produces lightweight time-step-wise gating weights; for low-dimensional ego states, it generates a fine-grained feature-time matrix.

Let $Q^{\text{uncer}} \in \mathbb{R}^{N \times d_h}$ denote the current uncertainty-aware queries, where $N = N_s + N_d$. When applying the gate to temporal perception queries $Q_\text{temporal} \in \mathbb{R}^{N_{tq} \times T \times d_h}$ (where $N_{tq}$ is large), it generates a vector $\mathbf{g}^{(t)} \in [0,1]^T$ that weights each time step uniformly across all queries and features. First, the $N$ uncertainty queries are aggregated into a global context $\mathbf{c} \in \mathbb{R}^{d_h}$ via attention:
\begin{equation}
\mathbf{c} = \sum_{i=1}^{N} \alpha_i \mathbf{q}_i, \quad \alpha_i = \frac{\exp(f_{\text{attn}}(\mathbf{q}_i))}{\sum_{j=1}^{N} \exp(f_{\text{attn}}(\mathbf{q}_j))},
\label{eq:gate_context}
\end{equation}
where $\mathbf{q}_i$ is the $i$-th row of $Q^{\text{uncer}}$ and $f_{\text{attn}}: \mathbb{R}^{d_h} \to \mathbb{R}$ is a linear layer. This context is then projected to $T$ dimensions and passed through a sigmoid:
\begin{equation}
\mathbf{g}^{(t)} = \sigma(\, f_{\text{gate}}^{(t)}(\mathbf{c}) \,),
\label{eq:gating_vec_temp}
\end{equation}
with $f_{\text{gate}}^{(t)}: \mathbb{R}^{d_h} \to \mathbb{R}^T$. The gating vector is broadcast to shape $[1, T, 1]$ and applied element-wise:
\begin{equation}
Q_\text{temporal}^\text{gated} = \mathbf{g}^{(t)} \odot Q_\text{temporal}.
\label{eq:gate_temp}
\end{equation}
This lightweight design efficiently modulates all $N_{tq}$ queries per time step using a single weight $g^{(t)}_t$. The modulated features $Q_\text{temporal}^\text{gated}$ are then passed to the original planner design, enabling dynamic, uncertainty-aware filtering of historical temporal information.

The same gating mechanism can be adapted to historical ego states $S_\text{ego} \in \mathbb{R}^{L \times T}$ (where $L$ is the number of ego state features, e.g., turn signals, velocities, accelerations) by increasing the granularity of the gating signal. Instead of generating a vector $\mathbf{g}^{(t)} \in [0,1]^T$, we produce a matrix $\mathbf{g}^{(e)} \in [0,1]^{L \times T}$ using a separate projection layer $f_{\text{gate}}^{(e)}: \mathbb{R}^{d_h} \to \mathbb{R}^{L \times T}$ after the same context $\mathbf{c}$ (Eq.~\eqref{eq:gate_context}):
\begin{equation}
\mathbf{g}^{(e)} = \sigma(\, f_{\text{gate}}^{(e)}(\mathbf{c}) \,).
\end{equation}
This allows independent gating for each ego-state feature at every time step. The gated output is then computed as $S_\text{ego}^\text{gated} = \mathbf{g}^{(e)} \odot S_\text{ego}$, and replaces the original ego status input to the planner, keeping the rest of the model architecture unchanged.

\section{Experiments}
We evaluate UniUncer on nuScenes and NavsimV2, showing improved trajectory accuracy ($7\%$ lower L2 error) and driving score ($10.8\%$ higher EPDMS) over strong baselines. Ablation studies confirm that modeling uncertainty in both static and dynamic elements, combined with the uncertainty-aware gate, consistently enhances performance.

\subsection{Main Results}

\textbf{Dataset.} We use nuScenes~\cite{caesar2020nuscenes} to evaluate our open-loop performance. The nuScenes dataset contains 1000 driving scenes recorded with six cameras and a 32-beam LiDAR sensor. It provides labels for dynamic objects, static map elements, and the trajectories of both the ego vehicle and dynamic objects, which are used for planning and prediction tasks. For pseudo closed-loop evaluation, we adopt NavsimV2~\cite{cao2025pseudo}, a planning-oriented autonomous driving dataset built on top of OpenScene, which itself is a redistribution of nuPlan. Navsim~\cite{dauner2024navsim} includes eight 1920×1080 cameras and a fused LiDAR point cloud. It was originally designed for open-loop evaluation of E2E methods. Later, NavsimV2 extended this work by rendering novel views with 3DGS to construct a pseudo closed-loop simulation.

\textbf{Evaluation Metrics.} For nuScenes, we report the L2 displacement error of the ego vehicle’s future trajectory and the collision rate. For NavsimV2, we adopt the Extended Predictive Driver Model Score (EPDMS)~\cite{cao2025pseudo} as the pseudo closed-loop planning evaluation metric, particularly on its most challenging benchmark, the Navhard two-stage test. EPDMS is a weighted sum of several sub-metrics, including No At-Fault Collisions (NC), Drivable Area Compliance (DAC), Driving Direction Compliance (DDC), Lane Keeping (LK), Time-to-Collision (TTC), History Comfort (HC), Extended Comfort (EC), Traffic Light Compliance (TLC), and Ego Progress (EP), providing a comprehensive closed-loop planning score. The specific EPDMS is defined as:
\begin{equation}
\begin{aligned}
\mathrm{EPDMS} = {} &
\prod_{m \in \mathcal{M}_{\mathrm{pen}}} 
   \mathrm{filter}_m(\mathrm{agent}, \mathrm{human})
   \cdot \\
& \frac{\sum_{m \in \mathcal{M}_{\mathrm{avg}}} 
   w_m \,\mathrm{filter}_m(\mathrm{agent}, \mathrm{human})}
   {\sum_{m \in \mathcal{M}_{\mathrm{avg}}} w_m },
\end{aligned}
\label{eq:epdms}
\nonumber
\end{equation}
where $\mathcal{M}_{\mathrm{pen}} = \{\mathrm{NC}, \mathrm{DAC}, \mathrm{DDC}, \mathrm{TLC}\}$ denotes the penalty terms, and $\mathcal{M}_{\mathrm{avg}} = \{\mathrm{TTC}, \mathrm{EP}, \mathrm{HC}, \mathrm{LK}, \mathrm{EC}\}$ denotes weighted average terms.

\begin{table*}[htbp]
% \vspace{+5pt}
\caption{Ablation study on the effectiveness of dynamic uncertainty and the uncertainty-aware gate on the Navhard two-stage test.}
\scriptsize
\centering
% \rowcolors{2}{gray!30}{gray!15}
% \resizebox{\textwidth}{!} {
\resizebox{\linewidth}{!} {
\begin{tabular}{l|c|c| c |c c c c c c c c c c | c}

    \toprule
    Method
    & D-Uncer
    & Uncer-Gate
    & Stage
    & $\text{NC $\uparrow$}$
    & $\text{DAC $\uparrow$}$
    & $\text{DDC $\uparrow$}$
    & $\text{TLC $\uparrow$}$
    & $\text{EP $\uparrow$}$
    & $\text{TTC $\uparrow$}$
    & $\text{LK $\uparrow$}$ 
    & $\text{HC $\uparrow$}$ 
    & $\text{EC $\uparrow$}$
    & $\text{EPDMS Stage $\uparrow$}$
    & $\text{EPDMS $\uparrow$}$ \\
    \midrule

    \multirow{2}{*}{DiffusionDrive~\cite{liao2025diffusiondrive}} & & & Stage 1 &
    96.3 & 84.7 & 98.7 & 99.3 & 84.3 & 95.3 & 94.4 & 97.8 & 76.4 & 65.4 &
    \multirow{2}{*}{25.9} \\
    & & & Stage 2 & 
    79.5 & 71.2 & 84.9 & 98.3 & 86.5 & 75.6 & 46.9 & 96.2 & 70.9 & 39.6 & \\
    \midrule
    
    \multirow{2}{*}{+D-Uncer} & \multirow{2}{*}{\checkmark} & & Stage 1 & 
    95.9 & 85.6 & 98.7 & 99.3 & 84.5 & 95.1 & 94.9 & 97.8 & 76.0 & \textbf{65.5} &  \multirow{2}{*}{26.7} \\
    & & & Stage 2 &
    79.2 & 72.3 & 85.9 & 98.2 & 86.6 & 76.2 & 48.2 & 96.3 & 70.1 & 40.6 & \\
    \midrule

    \multirow{2}{*}{+D-Uncer +Uncer-Gater} & \multirow{2}{*}{\checkmark} & \multirow{2}{*}{\checkmark} & Stage 1 & 
    96.8 & 82.9 & 98.8 & 99.3 & 83.6 & 95.8 & 96.0 & 97.6 & 77.3 & 64.1 &  \multirow{2}{*}{\textbf{28.7}} \\
    & & & Stage 2 &
    81.5 & 74.4 & 86.9 & 86.9 & 85.2 & 78.6 & 49.2 & 95.7 & 74.2 & \textbf{43.4} & \\

\bottomrule
\end{tabular}}
\label{tab:navhard_ablation}
\vspace{-10pt}
\end{table*}

\textbf{Evaluated Models.} We implement our UniUncer on top of SparseDrive~\cite{sun2025sparsedrive} for open-loop evaluation on nuScenes. We also incorporate UniUncer into DiffusionDrive~\cite{liao2025diffusiondrive} for comparative pseudo close-loop analysis. The main model architectures remain largely unchanged, with three key modifications: (1) replacing the deterministic heads for static and dynamic objects with probabilistic heads that estimate both $\mu$ and $b$ associated with object vertices; (2) adding two \textit{Uncertainty Fusion Modules} to incorporate the estimated uncertainties into static and dynamic queries; and (3) introducing an \textit{Uncertainty-aware Gate} for gating temporal object-related features in SparseDrive for every time step. For DiffusionDrive, the \textit{Uncertainty-aware Gate} is instead applied to the historical ego status, as it does not include temporal object features.

\begin{table*}[htbp]
\vspace{+5pt}
\caption{Ablation study on the effectiveness of static uncertainty, dynamic uncertainty, and the uncertainty-aware gate on nuScenes.}
\scriptsize
\centering
\resizebox{\textwidth}{!} {
\begin{tabular}{l|c|c|c|c c c c|c c c c}
\toprule
Method 
& S-Uncer 
& D-Uncer 
& Uncer-Gate 
& \multicolumn{4}{c|}{L2 (m) $\downarrow$} 
& \multicolumn{4}{c}{Col. Rate (\%) $\downarrow$} \\
\cline{5-12}
& & & & 1s & 2s & 3s & Avg. & 1s & 2s & 3s & Avg. \\
\midrule
SparseDrive~\cite{sun2025sparsedrive} & & & & 
0.29& 0.58 & 0.96 & 0.61 & 0.01 & 0.05 & 0.18 & 0.08 \\
\midrule
+S-Uncer & \checkmark & & & 
0.30 & 0.58 & 0.95 & 0.61 & 0.01 & 0.05 & 0.18 & 0.08 \\
\midrule
+D-Uncer & & \checkmark & & 
\textbf{0.27} & \textbf{0.53} & 0.90 & 0.57 & 0.02 & 0.11 & 0.21 & 0.11 \\
\midrule
+S-Uncer +D-Uncer & \checkmark & \checkmark & & 
\textbf{0.27} & \textbf{0.53} & \textbf{0.88} & \textbf{0.56} & 0.01 & 0.06 & 0.19 & 0.09 \\
\midrule
+S-Uncer +D-Uncer +Uncer-Gate & \checkmark & \checkmark & \checkmark & 
\textbf{0.27} & \textbf{0.53} & 0.90 & 0.57 & \textbf{0.00} & \textbf{0.04} & \textbf{0.17} & \textbf{0.07} \\
\bottomrule
\end{tabular}}
\label{tab:nuscenes_ablation}
\vspace{-10pt}
\end{table*}

\textbf{Evaluation on nuScenes.} As shown in Tab.\ref{tab:nuscenes_results}, UniUncer improves the L2 trajectory displacement at all time steps, achieving a $7\%$ reduction on average. Moreover, it achieves a lower collision rate, average collision rate reduced from $0.08\%$ to $0.07\%$, compared to the baseline, demonstrating improved safety without sacrificing trajectory accuracy.

\textbf{Evaluation on NavsimV2.} As shown in Tab.\ref{tab:navhard_results}, our model improves the overall EPDMS by $10.8\%$, increasing from 25.9 to 28.7. The stage two EPDMS also improves significantly, from 39.6 to 43.4, while the stage one score experiences a slight decrease. We attribute this to the fact that stage two contains many synthetically generated challenging scenarios with artifacts that can confuse perception, where uncertainty estimation provides substantial benefits. This result highlights the effectiveness of our design in handling complex and safety-critical driving conditions where perception reliability is compromised. 

\textbf{Training Details.}
We keep the data format identical to the baselines, only modifying dynamic object labels by replacing box parameters with vertices. Four corner points and one center point serve as regression targets. The static and dynamic branches are supervised by a combination of $\ell_1$ and Negative Log-Likelihood (NLL) losses:
\[
L_{\text{static/dynamic}} = w_1 \cdot \ell_1^{\text{static/dynamic}} + w_2 \cdot \text{NLL}_{\text{static/dynamic}},
\]
where $w_1$ and $w_2$ balance the losses ($w_1=0.25$, $w_2=0.6$ for dynamic; $w_1=0$, $w_2=1.0$ for static).

For nuScenes, we fine-tune from the official SparseDrive stage one checkpoint for 10 epochs to ensure convergence of the new probability heads, followed by another 10 epochs for stage two. For NavsimV2, we train from scratch using the same configuration as DiffusionDrive, since NavsimV2 excludes LiDAR data while the released DiffusionDrive weights were trained with LiDAR. All experiments use four NVIDIA RTX 3090 GPUs.

\subsection{Ablation Study}
We conduct ablation studies on both NavsimV2 and nuScenes to validate the contribution of each component in UniUncer.

\textbf{Notation in the ablation tables.} In Tab.\ref{tab:navhard_ablation} and Tab.\ref{tab:nuscenes_ablation}, we use 'S-Uncer' and 'D-Uncer' to denote the uncertainty estimation for static and dynamic objects, respectively. 'Uncer-Gate' denotes the uncertainty-aware gate. Note that for the NavsimV2 ablation~\ref{tab:navhard_ablation}, we focus on dynamic uncertainty and the gate, as the dataset does not provide vectorized map labels required for training the static uncertainty head. This highlights a practical limitation but also showcases the modular benefit of our framework and components can be applied independently where data permits.

\textbf{Effectiveness of uncertainty estimation for dynamic objects and uncertainty-aware gate.} As shown in Tab.\ref{tab:navhard_ablation}, we report the performance on the Navhard two-stage evaluation using DiffusionDrive as the baseline and progressively adding our proposed modules. With uncertainty estimation and the uncertainty fusion module, the EPDMS improves from 25.9 to 26.7. Incorporating all modules, including the uncertainty-aware gate, further increases the EPDMS to 28.7. These steady improvements demonstrate the effectiveness of our proposed modules, validating both the contribution of uncertainty estimation for dynamic objects and the role of uncertainty-aware gate for past ego status. We do not involve static uncertainty, since the vectorized label for the map element is not available in NavsimV2.
% change the number accroding to tabel3

\textbf{Effectiveness of uncertainty estimation for dynamic and static objects and uncertainty-aware gate.} Tab.\ref{tab:nuscenes_ablation} presents an ablation study in which we gradually incorporate our proposed modules into SparseDrive on nuScenes data set. Introducing uncertainty for static objects alone provides limited improvement in L2 displacement, consistent with results reported in UncAD~\cite{yang2025uncad}. When uncertainty is applied to dynamic objects only, the average L2 displacement error decreases from 0.61 m to 0.57 m, although the collision rate slightly increases. Incorporating both static and dynamic uncertainty leads to improvements in both L2 error and collision rate, reducing the average L2 error to 0.56 m while the average collision rate only slightly increases from $0.08\%$ to $0.09\%$. Finally, when the full model, combining static and dynamic uncertainty estimation with the uncertainty-aware gate, we observe consistent gains in both metrics: the average L2 error decreases to 0.57 m, and the average collision rate drops to $0.07\%$. These results demonstrate that explicitly modeling both static and dynamic uncertainties, together with adaptive gating of historical features, effectively enhances trajectory prediction accuracy while reducing collisions, highlighting the practical benefit of our unified uncertainty framework for E2E autonomous driving.

\begin{figure}[t]
    \centering
    \includegraphics[width=0.45\textwidth]{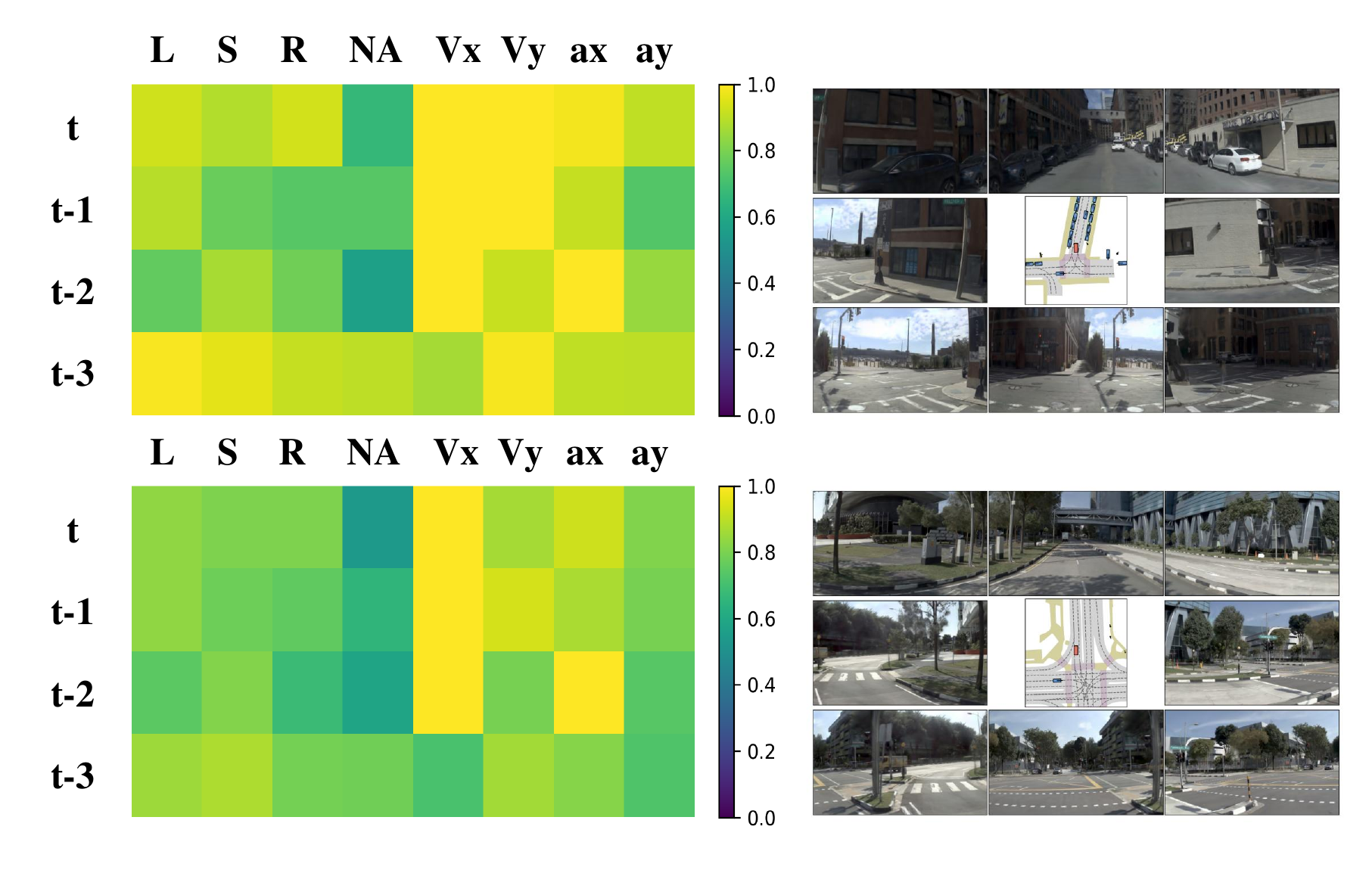}
    \vspace{-10pt}
    \caption{Visualization of the Uncertainty-aware Gate (fine-grained feature-
time matrix mode) for Ego Status.
    The upper row illustrates a complex scenario, while the lower row corresponds to a simple scenario. 
    Driving commands include \textit{L} (left turn), \textit{S} (keep straight), \textit{R} (right turn), and \textit{NA} (not available). 
    \textit{Vx} and \textit{Vy} denote velocities along the $x$- and $y$-axes, respectively, while \textit{ax} and \textit{ay} represent accelerations along the $x$- and $y$-axes.}
    \label{fig:uncertainty_gate}
% \vspace{-10pt}
\end{figure}

\textbf{Efficiency of Our Model.}
We compare our model with SparseDrive, which serves as the backbone of our method. As shown in Tab.\ref{tab:nuscenes_speed}, our model introduces only 5.6 M additional parameters compared to the baseline, while incurring just a 0.5 FPS reduction. This result aligns with our design goal: enabling unified uncertainty estimation with only minimal modifications to the original architecture.

\begin{table}[htbp]
\caption{Efficiency of Our Approach. All methods are tested on a single RTX 3090 GPU.}
\centering
\resizebox{0.8\linewidth}{!} {
\begin{tabular}{l |c | c}
\toprule
Method & Number of Params (M) & FPS \\
\midrule
SparseDrive~\cite{sun2025sparsedrive} & 86.2 & 6.6 \\
\midrule
SparseDrive + Ours & 91.8 & 6.1 \\
\bottomrule
\end{tabular}
}
\label{tab:nuscenes_speed}
\vspace{-10pt}
\end{table}

\subsection{Qualitative Results}

\begin{figure*}[htbp]
    \centering
    \includegraphics[width=0.8\linewidth]{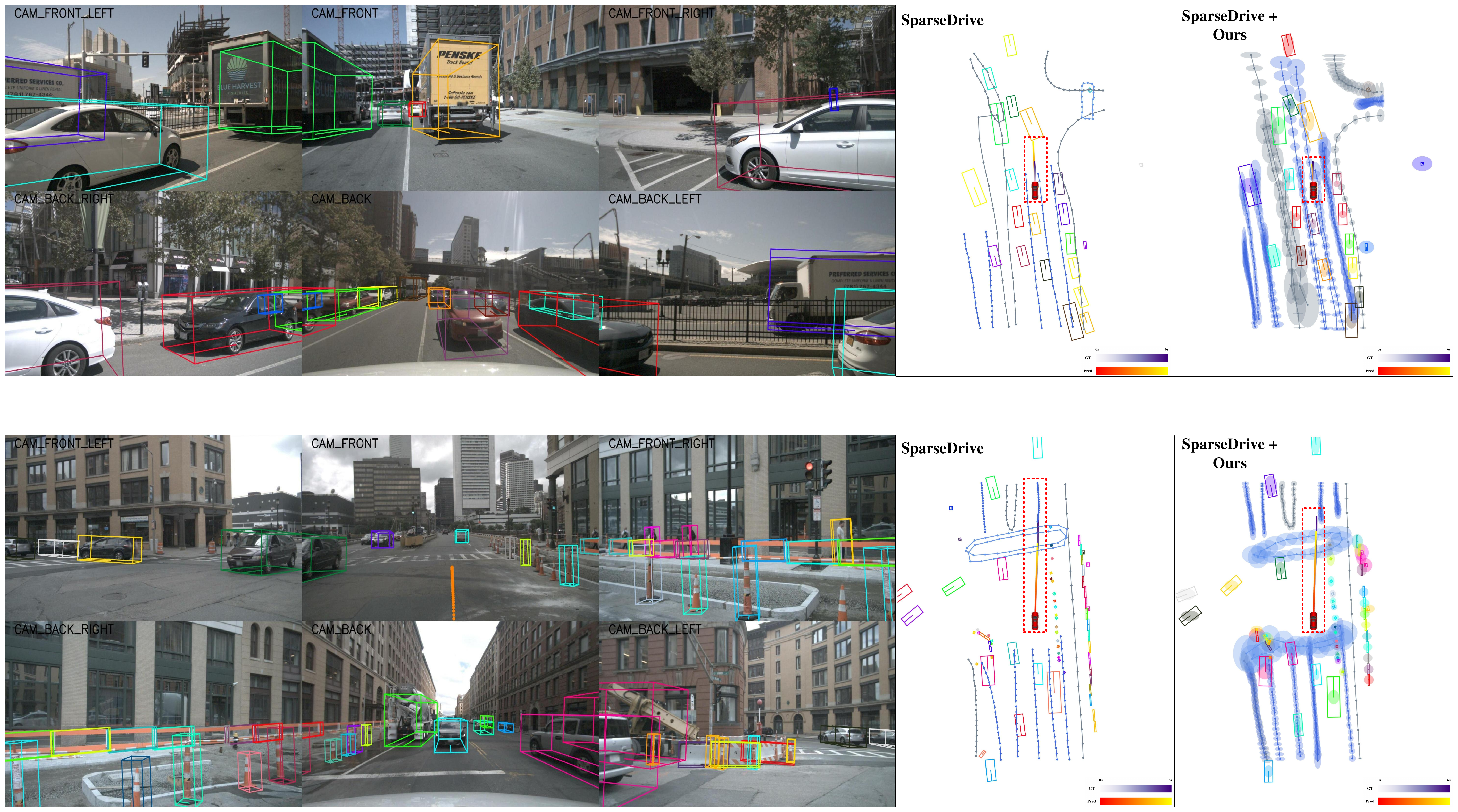}
    \vspace{-5pt}
    \caption{Qualitative results on the nuScenes dataset. With unified uncertainty estimation, the predicted trajectory from our model is closer to the ground-truth trajectory compared to the baseline model. Ellipses on map vertices and dynamic objects represent the scale parameter $b$ along the $x$- and $y$-axes. For dynamic objects, the uncertainty is lower on the visible side of the vehicle compared to the invisible side. Best viewed in the digital version.}
    \label{fig:nusc_vis}
\vspace{-5pt}
\end{figure*}

\begin{figure*}[htbp]
\centering
    \includegraphics[width=0.8\linewidth]{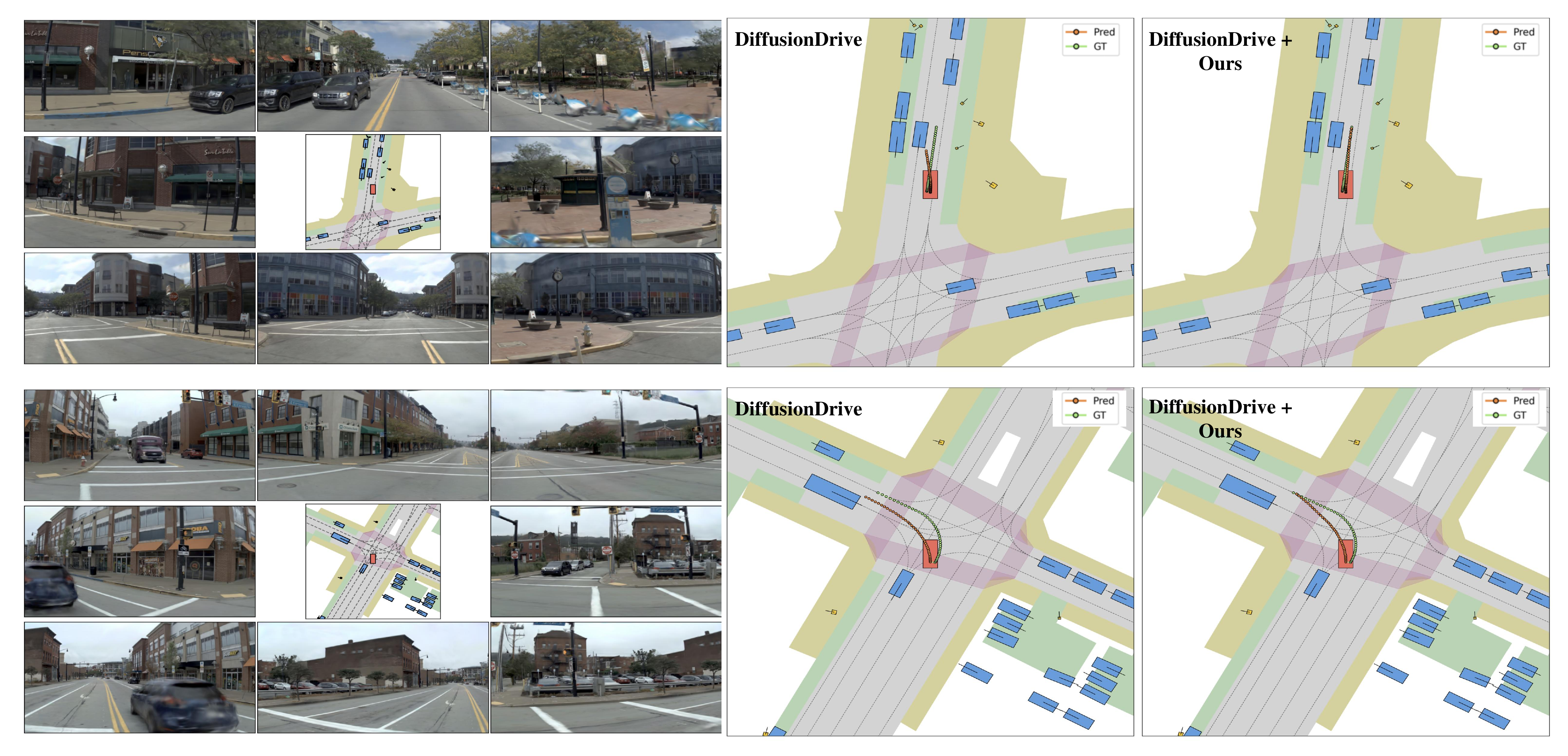}
    \vspace{-5pt}
    \caption{Qualitative results on the Navhard two-stage test. 
    Uncertainty estimation enables the planner to generate trajectories that are closer to the ground-truth (GT) trajectory in general. 
    It demonstrates that uncertainty estimation for dynamic agents is critical for collision avoidance in challenging pseudo closed-loop scenarios rendered by 3DGS, which can introduce synthetic artifacts (e.g., the bus and vehicle in the second row). By assigning higher uncertainty to such potentially unreliable detections, the planner adopts a more conservative strategy, successfully avoiding over-reaction and improving robustness. Better view in digital version.}
\label{fig:navsim_vis}
\vspace{-15pt}
\end{figure*}

We visualize model behavior to demonstrate the effectiveness of our modules.

Using ego status from $t$ to $t-3$ (commands, velocities, accelerations), Fig.~\ref{fig:uncertainty_gate} shows that the uncertainty-aware gate emphasizes richer historical information in complex multi-object scenes, while prioritizing $x$-axis velocity and acceleration in simpler scenes.

Fig.~\ref{fig:nusc_vis} displays estimated uncertainties and planned trajectories on nuScenes. Well-calibrated $x$- and $y$-axis uncertainties help the planner produce trajectories closer to human demonstrations, indicating more natural and robust behavior. (Uncertainties per vehicle are aggregated at its center for visualization.)

Fig.~\ref{fig:navsim_vis} illustrates that dynamic uncertainty is critical for collision avoidance in challenging 3DGS-rendered scenes, where synthetic objects may exhibit artifacts. By assigning higher uncertainty to unreliable detections, the planner adopts a conservative strategy, avoiding over-reaction and improving safety.

\section{Conclusion}

We present UniUncer, a lightweight and plug-and-play framework that models uncertainty in both dynamic and static objects for end-to-end driving. It leverages this unified uncertainty to adaptively gate historical information such as ego states and temporal queries. Experiments on nuScenes and NavsimV2 show consistent improvements in open-loop and pseudo closed-loop evaluations, enhancing both trajectory accuracy and safety. The design adds minimal overhead and is readily adaptable to various E2E backbones. This work encourages future research on interactive uncertainty and epistemic uncertainty estimation for out-of-distribution robustness.

% \newpage
\clearpage

\bibliographystyle{IEEETran}
\bibliography{main}

\end{document}